\def\year{2026}\relax

\documentclass[letterpaper]{article} 

\usepackage[submission]{aaai2026}  
\usepackage{times}  
\usepackage{helvet}  
\usepackage{courier}  
\usepackage[hyphens]{url}  
\usepackage{graphicx} 
\usepackage{natbib}  
\usepackage{caption} 
\usepackage{algorithm}
\usepackage{algorithmic}
\usepackage{amssymb}
\usepackage{amsmath}
\usepackage{booktabs}
\usepackage{multirow}
\usepackage{xcolor}
\usepackage{colortbl}
\usepackage{makecell}

\usepackage{newfloat}
\usepackage{listings}

\makeatletter
\let\showauthors@on=T  
\makeatother

\DeclareCaptionStyle{ruled}{labelfont=normalfont,labelsep=colon,strut=off} 
\floatstyle{ruled}
\newfloat{listing}{tb}{lst}{}
\floatname{listing}{Listing}
\title{QueueEDIT: Structural Self-Correction for Sequential Model Editing in LLMs}

\author{Taolin Zhang$^{1}$, Haidong Kang$^{2}$, Dongyang Li$^{4}$, Qizhou Chen$^{5}$, Chengyu Wang$^{3}$\thanks{\ \ Co-corresponding authors} \\ Xiaofeng He $^{5}$, Richang Hong $^{1}$\footnotemark[1]}
\affiliations {
$^1$ School of Computer Science and Information Engineering, Hefei University of Technology \\
$^2$ College of Software, Northeastern University
$^3$ Alibaba Cloud Computing\\
$^4$ Shanghai University of Electric Power
$^5$ East China Normal University \\
 {tlzhang@hfut.edu.cn, chengyu.wcy@alibaba-inc.com}
}

\usepackage{bibentry}

\begin{document}

\maketitle

\begin{abstract}
Recently, large language models (LLMs) have demonstrated impressive results but still suffer from hallucinations. Model editing has been proposed to correct factual inaccuracies in LLMs. A challenging case is sequential model editing (SME), which aims to rectify errors continuously rather than treating them as a one-time task. During SME, the general capabilities of LLMs can be negatively affected due to the introduction of new parameters.
In this paper, we propose a queue-based self-correction framework (\texttt{QueueEDIT}) that not only enhances SME performance by addressing long-sequence dependency but also mitigates the impact of parameter bias on the general capabilities of LLMs. Specifically, we first introduce a structural mapping editing loss to map the triplets to the knowledge-sensitive neurons within the Transformer layers of LLMs.
We then store the located parameters for each piece of edited knowledge in a queue and dynamically align previously edited parameters. In each edit, we select queue parameters most relevant to the currently located parameters to determine whether previous knowledge needs realignment. Irrelevant parameters in the queue are frozen, and we update the parameters at the queue head to the LLM to ensure they do not harm general abilities.
Experiments show that our framework significantly outperforms strong baselines across various SME settings and maintains competitiveness in single-turn editing. The resulting LLMs also preserve high capabilities in general NLP tasks throughout the SME process.~\footnote{Source code will be released upon paper acceptance.}
\end{abstract}

\section{Introduction}
Recently, large language models (LLMs) have become the foundational infrastructure of modern NLP~\cite{DBLP:conf/acl/ZhengZCHTL0M22,DBLP:conf/acl/BlinovaZJEB23}. However, LLMs occasionally generate undesirable outputs~\cite{DBLP:journals/nca/BastaCC21,DBLP:journals/corr/abs-2310-20689} and tend to produce hallucinations~\cite{DBLP:conf/icml/ShiCMSDCSZ23,DBLP:conf/acl/TamMZKBR23}, creating content that appears plausible but lacks factual support. Although fine-tuning models with updated knowledge offers a direct solution, it is often impractical due to excessive time requirements~\cite{DBLP:conf/iclr/HubotterBH025,kim2025plugin,krishna2025}. To address these issues, there is increasing interest in incorporating knowledge into LLMs via model editing, which directly adjusts a small subset of parameters~\cite{DBLP:conf/emnlp/MadaanTCY22,DBLP:conf/iclr/FangJWMSW0C25}.

\begin{figure*}[!t]
  \centering
  \includegraphics[width=16cm]{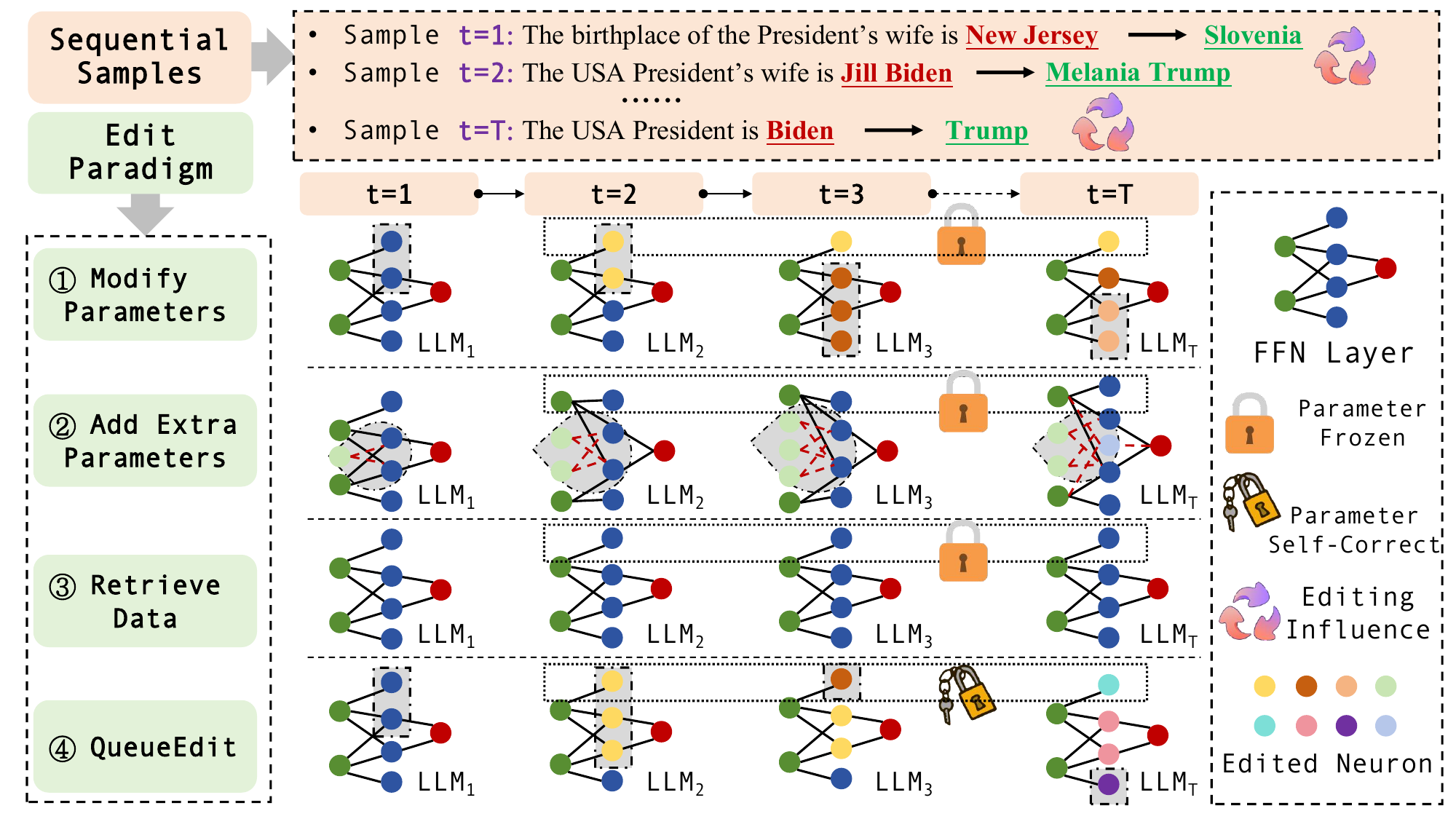}
\caption{Comparison between SME methods. 
\texttt{Modifying Parameters} and \texttt{Adding Extra Parameters} approaches change parameters at specific positions in FFN layers without making corresponding adjustments for relevant edits. \texttt{Retrieving Data} methods focus solely on external data for editing knowledge triples. \texttt{QueueEDIT} not only dynamically updates parameters associated with previous edits but also ensures the general capabilities of LLMs.}
    \vspace{-0.5cm}
  \label{motivation_res}
\end{figure*}

In the literature, previous approaches broadly fall into three categories: \texttt{Modifying Parameters}, \texttt{Adding Extra Parameters}, and \texttt{Retrieving Data}.  
(1) \texttt{Modifying Parameters} approaches modify one or a batch of knowledge triples at a time by editing LLM parameters using meta-learning~\cite{KnowledgeEditor,DBLP:conf/iclr/MitchellLBFM22} or locate-then-edit techniques~\cite{GRACE,DBLP:conf/nips/MengBAB22,DBLP:conf/iclr/MengSABB23}. These methods involve locating the neurons in FFN layers corresponding to the edited data and updating these parameters.  
(2) \texttt{Adding Extra Parameters} methods augment LLMs with supplementary neurons for each edit, thus bypassing alterations to the original parameters~\cite{DBLP:conf/icml/MitchellLBMF22,T-Patcher,DBLP:conf/acl/ZhangCL0HHXH24}. Here, parameter updates are not achieved through backpropagation but rather through extra modules associated with each fact. While these methods quickly fix mistakes, previously edited parameters relevant to the current edit are not updated simultaneously.  
(3) \texttt{Retrieving Data} approaches retrieve external information related to knowledge triplets and concatenate it into the input to form the final editing data~\cite{han-etal-2023-improving,LTE,DBLP:conf/emnlp/ChenZHL0HX24}. However, continuous retrieval and concatenation can lead to a rapid increase in the length of LLM input, resulting in longer inference time~\cite{DBLP:conf/emnlp/0001DGL23,DBLP:journals/corr/abs-2409-10516}.  
Additionally, as the number of edits increases, outdated editing information can cause incorrect answers, negatively affecting LLM performance due to knowledge bias. In Figure~\ref{motivation_res}, when the current editing instance ``Donald Trump'' is ready to be updated, parameters related to previous edits
should be adjusted accordingly. Otherwise, the subsequent question about the ``USA President’s wife'' will still yield the incorrect answer ``Jill Biden.''

In this paper, we propose a queue-based self-correction framework named \texttt{QueueEDIT}. This framework performs knowledge editing for each triplet while simultaneously adjusting specific parameters in FFNs that were previously updated, aiming to maintain SME performance and preserve the general capabilities of LLMs. The key techniques of \texttt{QueueEDIT} are introduced as follows:

\noindent\textbf{Structural Mapping Editing.} To more accurately update the knowledge of the current edit to specific FFN layers~\cite{DBLP:conf/nips/MengBAB22,DBLP:conf/iclr/FangJWMSW0C25}, we map the triplets (i.e., \(<s,r,o>\)\footnote{The editing samples are sourced from knowledge graph (KG) triples \(<subject, relation, object>\) ($<s,r,o>$).}) to knowledge-sensitive positions in FFN layers. This approach enhances the response of internal activation parameters compared to previous works that disregarded the structural semantics of knowledge triplets~\cite{DBLP:conf/iclr/MitchellLBFM22,DBLP:journals/corr/abs-2401-06855}. Previous methods utilized both the ``relation'' and ``object'' to learn representations for gradient feedback, overlooking the original structure of the knowledge triplets. In this paper, we map the subject, relation, and object in the editing data to the first MLP matrix, the second MLP matrix, and the gradient backpropagation representation in the Transformer’s FFN layer, respectively. Specifically, the ``relation'', originally viewed as a bridge connecting the semantic knowledge of subject and object entities~\cite{DBLP:conf/nips/BordesUGWY13,DBLP:conf/aaai/WangZFC14}, is independently mapped to the second parameter matrix in the FFN layer. We then design a structural loss to update the parameters via the triple’s representations, enhancing editing performance.

\noindent\textbf{Queue-based Self-Correction.} In the context of SME, it is crucial to consider the inherent semantic associations among edited data. To manage dependency relationships in a long sequence of edits, we design a queue-based structure to store updated editing parameters following the first-in, first-out (FIFO) principle. Specifically, we insert the current editing parameters at the end of the queue and calculate the distance between the current parameters and those already in the queue. After sorting by distance, we update the original parameters in the queue by comparing the gradient representation of the object entity with the relation representation of the current edit. For computational efficiency, we select only the top-\(K\) editing parameters for sequential updates. Finally, we remove the oldest parameters from the queue head to ensure the freshness of knowledge.

In our experiments, we evaluate \texttt{QueueEDIT} against model editing baselines on both single-turn and multi-turn editing, as well as on general NLP datasets. For SME, our method significantly surpasses baselines as the number of edits increases and even shows a slight advantage in single-turn editing. Regarding the analysis of general capabilities, our model shows better consistency than the original LLMs without any degradation in overall LLM performance.


\section{Related Works on LLM Editing}

\noindent\textbf{Modifying Parameters.} These approaches can be further divided into Locate-then-Edit and meta-learning-based methods. For Locate-then-Edit, ROME~\cite{DBLP:conf/nips/MengBAB22} and MEMIT~\cite{DBLP:journals/corr/abs-2311-04661} introduce a causal intervention framework to identify neuron activations and then update the knowledge data with low rank-based model editing to modify parameters. AlphaEdit~\cite{DBLP:conf/iclr/FangJWMSW0C25} projects perturbations onto the null space of preserved knowledge prior to their application to the model parameters.
For meta-learning methods, KnowledgeEditor~\cite{KnowledgeEditor} and MEND~\cite{DBLP:conf/iclr/MitchellLBFM22} employ distinct methodologies, converting edited knowledge and decomposed gradients of the LLM into weight offset modifications, respectively. MALMEN~\cite{DBLP:journals/corr/abs-2311-04661} advances this paradigm further by incorporating normal equations to optimize parameter integration for batch editing operations. DAFNet~\cite{DBLP:conf/acl/ZhangCL0HHXH24} introduces intra-editing and inter-editing attention flows to update weighted representations at the sequence-level granularity.
While these methods demonstrate efficacy in single or batch editing, the progressive accumulation of parameter modifications with an increasing number of edits may ultimately lead to editing degradation~\cite{WILKE}.

\noindent\textbf{Adding Extra Parameters.} This paradigm does not directly modify specific FFN layer parameters but adds training modules at corresponding positions to incorporate the edited data into the LLM. CaLiNet~\cite{CALINET} and T-Patcher~\cite{T-Patcher} achieve model editing by introducing additional neurons to the LLM for each piece of editing knowledge, thereby avoiding modifications to the original model parameters. 
GRACE~\cite{GRACE} employs an adapter module that establishes a mapping between input queries and their corresponding knowledge representations.
However, in an SME scenario, the persistent incorporation of new neurons to update the current edited data can overlook the correlations between different sequential editing data and increase the burden on model inference speed.

\noindent\textbf{Retrieving Data.} This approach enhances the model's reasoning capabilities in a timely manner by continuously retrieving external data to supplement editing data without introducing additional parameters. Extending the GRACE framework, MELO~\cite{MELO} proposes a batch editing implementation leveraging LoRA technology~\cite{DBLP:conf/iclr/HuSWALWWC22}.
LTE~\cite{LTE} fine-tunes the LLM to generate appropriate responses when provided with knowledge prefixed by editing cues, while leveraging the pre-trained backbone architecture for relevant content retrieval~\cite{DBLP:conf/emnlp/ReimersG19}. However, the aforementioned methods, whether modifying parameters or retrieving external data, do not model the associations between sequential editing data in SME scenarios, resulting in logical errors between factual answers provided by LLMs.

\begin{figure*}[!t]
  \centering
  \includegraphics[width=17.5cm]{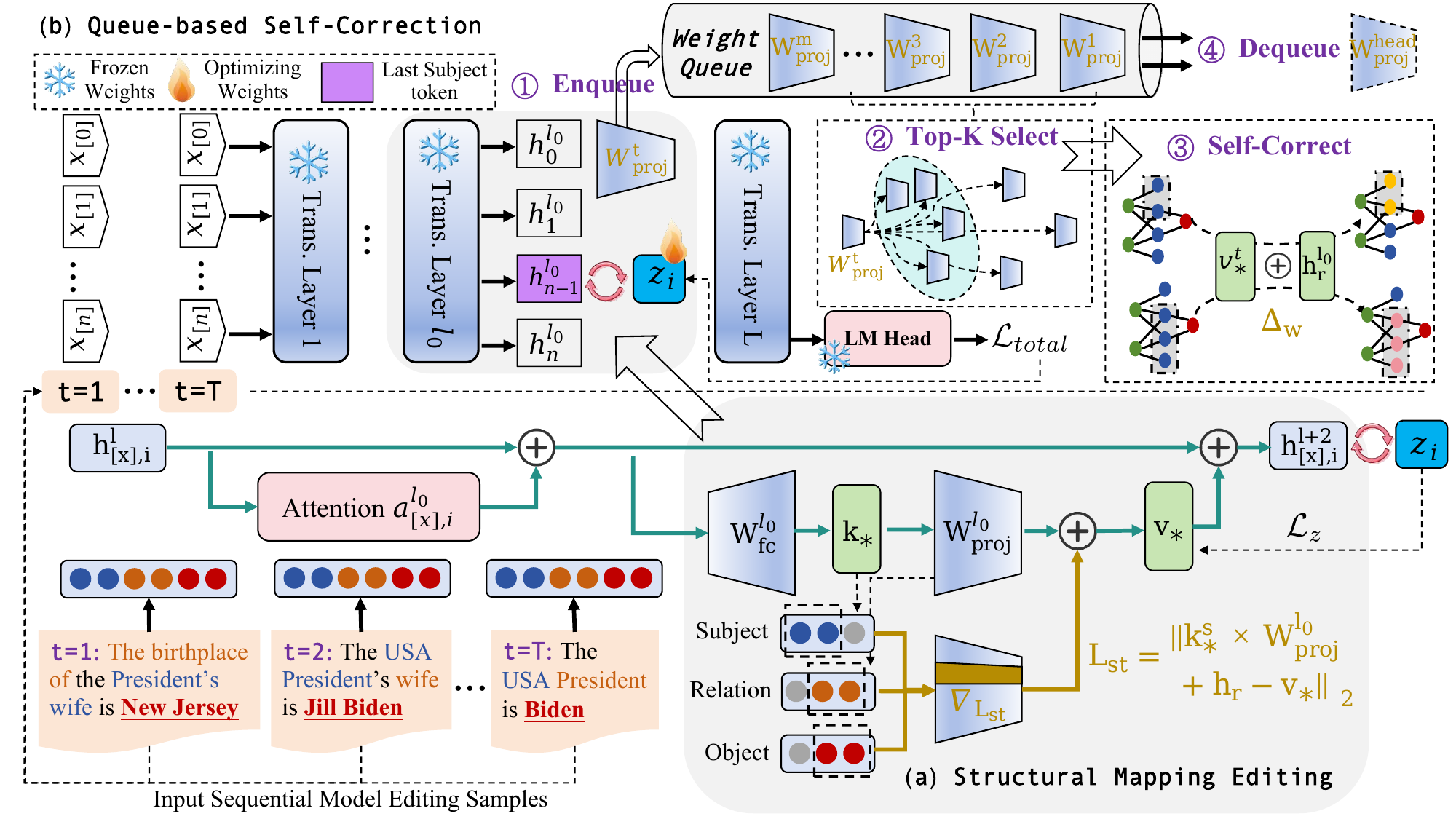}
    \caption{
    The \texttt{QueueEDIT} framework. 
    \textbf{(a) Structural Mapping Editing} maps each triple \((s,r,o)\) to dedicated positions (i.e., \(k_*^s\), \(h_r\), and \(v_*\)) within the FFN layer. 
    \textbf{(b) Queue-based Self-Correction} stores the located parameters into a \texttt{Weight Queue} following the FIFO principle for subsequent semantic alignment, aiming to preserve the general capabilities of LLMs. 
    }
  \label{model_overview}
      \vspace{-0.5cm}
\end{figure*}

\section{Methodology}
In this section, we introduce the basics of SME, including its definition, properties, and training losses. Then, we describe the main components of \texttt{QueueEDIT}, namely \textbf{Structural Mapping Editing} and \textbf{Queue-based Self-Correction}. The overall framework is illustrated in Figure~\ref{model_overview}.

\subsection{Preliminaries of SME Task}
\label{preliminary_task}
\noindent\textbf{Task Definition.}
Given an LLM $f$ and an edit example $(x_e, y_e)$ such that $f(x_e) \neq y_e$, a model editor (ME) outputs a post-edit model: $f' = \mathrm{ME}(f, x_e, y_e)$.
In SME, given a sequence of facts $\{(x_{e_1}, y_{e_1}), \ldots, (x_{e_T}, y_{e_T})\}$ and an initial model $f$, a ME needs to conduct edits successively:
$f_t = \operatorname{ME}\left(f_{t-1}, x_{e_t}, y_{e_t}\right)$
where $t = 1, \ldots, T$ and $f_0 = f$.
\noindent\textbf{Task Properties.} Every edit should satisfy basic properties: reliability, generality, and locality \cite{DBLP:conf/acl/ZhangCL0HHXH24}.
 
\noindent\textbf{Reliability:} A reliable edit holds when the post-edit model $f_t$ gives the target answer for the every cases $(x_{e_\tau}, y_{e_\tau}),\tau\leq t$ to be edited. The reliability is measured as the average accuracy on the edit cases:
\begin{equation}
\small
        \mathbb{E}_{\left(x_e, y_e\right) \sim\left\{\left(x_{e_\tau}, y_{e_\tau}\right)\right\}_{\tau=1}^t} \mathbb{I}\left\{\mathop{\operatorname{argmax}}\limits_y f_t\left(y \mid x_e\right)=y_e\right\}.
\end{equation}

\noindent\textbf{Generality:} The post-edit model $f_t$ should also satisfy the relevant neighbors $N(x_{e_\tau}, y_{e_\tau}), \tau\leq t$.
    It is evaluated by the average accuracy of $f_t$ on examples drawn uniformly from the relevant neighborhood:
    \begin{equation}
    \small
    \begin{split}
        \mathbb{E}_{\left(x_e, y_e\right) \sim\left\{\left(x_{e_\tau}, y_{e_\tau}\right)\right\}_{\tau=1}^t}\mathbb{E}_{(x_g, y_g) \sim N\left(x_{e}, y_{e}\right)} \operatorname{G}(x_g, y_g)\\
        \text{s.t.}\;\operatorname{G}(x_g, y_g) = \mathbb{I}\left\{\mathop{\operatorname{argmax}}\limits_y f_t \left(y \mid x_g\right)=y_g\right\}.
    \end{split}
    \end{equation}
    
\noindent\textbf{Locality:} Editing should be local, which means the post-edit model $f_t$ should not change the output of irrelevant examples in $O(x_{e_\tau}, y_{e_\tau}), \tau\leq t$. Hence, the locality is evaluated by the rate at which the post-edit model $f_t$’s predictions are unchanged as the pre-edit model $f$:
    \begin{equation}
    \small
    \begin{split}
        \mathbb{E}_{\left(x_e, y_e\right) \sim\left\{\left(x_{e_\tau}, y_{e_\tau}\right)\right\}_{\tau=1}^t} \mathbb{E}_{(x_l, y_l) \sim O(x_{e}, y_{e})} \operatorname{L}(x_l, y_l)\\
        \text{s.t.}\;\operatorname{L}(x_l, y_l) = \mathbb{I}\left\{f_t \left(y \mid x_l\right)=f\left(y \mid x_l\right)\right\}.
    \end{split}
    \end{equation}

\noindent\textbf{Task Training.} Let $(x_e^{(t)}, y_e^{(t)})$ be the reliability sample of the $t$-{th} fact, i.e., the editing sample itself.
$(x_{g_j}^{(t)}, y_{g_j}^{(t)})$ and 
$N_{g}^{(t)}$ and $N_{l}^{(t)}$ are the corresponding sample numbers of the $t$-{th} fact.
The total loss $\mathcal{L}_{ed}(f_T)$ is the sum of the following:
\begin{equation}
\small
    \mathcal{L}_{rel}(f_T) = \sum_{t=1}^T -\log{f_T(y_e^{(t)}|x_e^{(t)})},
\end{equation}
\begin{equation}
\small
\mathcal{L}_{gen}(f_T) = \sum_{t=1}^T\sum_{j=1}^{N_{g}^{(t)}} -\log{f_T(y_{g_j}^{(t)}|x_{g_j}^{(t)})},
\end{equation}
\begin{equation}
\small
\mathcal{L}_{loc}(f,f_T) = \sum_{t=1}^T \sum_{j = 1}^{N_{l}^{(t)}}\mathrm{KL}(f(x_{l_j}^{(t)})||f_T(x_{l_j}^{(t)})).
\end{equation}

\subsection{Structural Mapping Editing}
Our structural mapping editing approach maps the triplet structure to specific positions in the Transformer’s MLP layer, establishing semantic associations between the masked entity (i.e., the ``Object'') and the ``Subject'' entity through independent ``Relation'' learning.

\noindent\textbf{Locating the MLP Layer.}  
Following~\citet{DBLP:conf/nips/MengBAB22,DBLP:conf/iclr/MengSABB23,DBLP:conf/acl/ZhangCL0HHXH24}, MLP layers in Transformers are selected for editing via causal tracing. These studies analyze all internal activations through three experimental runs in a Transformer language model to identify the layer with the largest indirect effect, denoted as \(l_0\). The layers act as two-layer key–value memories, where neurons of the first layer, \(W_{fc}^{l_0}\), form a \emph{key}, and neurons of the second layer, \(W_{proj}^{l_0}\), represent the associated \emph{value}.

\noindent\textbf{Structural Triplet Editing.}  
Editing a sample \((x_e, y_e)\) constructed from a new triple \(t^* = (s,r,o^*)\) in place of \(t = (s,r,o)\) demonstrates a fine-grained understanding of the association-storage mechanisms.  
The parameters are derived via a closed-form solution~\cite{DBLP:conf/nips/MengBAB22,DBLP:conf/iclr/MengSABB23}:
\begin{equation}
\label{equ_7}
    W_{proj}^{l_0'} = W_{proj}^{l_0} + \Lambda (C^{-1} k_*)^T
\end{equation}
where \(\Lambda = \frac{(v_* - W_{proj}^{l_0} k_*)}{(C^{-1} k_*)^T k_*}\) and \(C = K \cdot K^T\) is a constant precomputed using cached Wikipedia embeddings \(K\) from the editing triples.\footnote{We use backpropagation considering the object entity \(o^*\) to obtain \(v_*\)~\cite{DBLP:conf/nips/MengBAB22}; please refer to Appendix~\ref{kv_calculation}.}  
Here, \(k_*\) is obtained by forwarding an average over \(N\) sampled texts \(x\) formed by templates consisting of the subject \(s\) and relation \(r\):
\begin{gather}
\label{eq_8}
    k_* = \frac{1}{N} \sum_{i=1}^{N} \sigma \left( W_{fc}^{l_0} \; \gamma\bigl(a_{[x],i}^{l_0} + h_{[x],i}^{(l_0 - 1)} \bigr) \right)
\end{gather}
where \(a_{[x],i}^{l_0}\) is the self-attention output at layer \(l_0\) and \(h_{[x],i}^{(l_0 - 1)}\) is the hidden input at layer \((l_0 - 1)\).  
\(\gamma\) is a normalization nonlinearity and \(\sigma\) is an activation function.  
From Eq.~\ref{eq_8}, we note that the relation \(r\) is mixed into the input \(x\) without separately modeling its structural semantics with the subject entity \(s\) and object entity \(o^*\).

Triples \(t^* = (s,r,o^*)\) can be represented by translation-based models~\cite{DBLP:conf/nips/BordesUGWY13}. These models imply that the semantics of object entities \(o^*\) can be approximated by subject entities \(s\) and relations \(r\). Here, we inject structural semantics into the parameters to enhance knowledge memorization in MLP layers. Specifically, we first leverage the subject entity \(s\) and relation \(r\) to generate \(k_*\) via the MLP parameters \(W_{fc}\) using Eq.~\ref{eq_8}. The hidden representations are:
\begin{gather}
    k_*^s = k_*[s_m : s_n], \quad k_*^r = k_*[r_m : r_n] \\
    h_r = \sigma\bigl(k_*^r W_{proj}^{l_0} + b_r\bigr)
\end{gather}
where \(s_m\) and \(s_n\) denote the subject token positions in the editing data, while \(r_m\) and \(r_n\) correspond to the relation positions. \(b_r\) is a bias term. \(k_*^s\) and \(h_r\) represent the subject and relation embeddings, respectively.

Next, we separate the relation \(r\) in triples and explicitly model it during the editing of the new knowledge triple \((s,r,o^*)\), rather than binding it with the subject entity \(s\) to transfer the contextual representation. The structural editing loss \(\mathcal{L}_{st}\) is defined as:
\begin{gather}
    \mathcal{L}_{st} = \big\| k_*^s W_{proj}^{l_0} + h_r - v_* \big\|_2
\end{gather}
where the gradient of \(\mathcal{L}_{st}\) is used only to optimize \(W_{proj}^{l_0}\), with all other parameters frozen.

By employing this translation-based loss, we constrain the transferred representation ``\((s + r) \rightarrow o^*\)'' to approximate the new knowledge \(o^*\) as closely as possible, while preserving the feature information of the old knowledge \(o\) stored in the located FFN layer \(W_{proj}^{l_0}\).

\subsection{Queue-Based Self-Correction}
\label{queue_self_correct}
In previous SME methods~\cite{KnowledgeEditor, DBLP:conf/iclr/MitchellLBFM22, DBLP:conf/acl/ZhangCL0HHXH24}, \(W_{proj}^{l_0}\) is repeatedly located for different triples, and knowledge is edited independently. Thus, the semantic influence among editing parameters is ignored, which leads to bias and adversely affects the general capabilities of LLMs~\cite{you2024shiftaddllm}. In \texttt{QueueEDIT}, we introduce a self-correction mechanism with a queue to update the located knowledge parameters in each edit. Specifically, after each edited parameter is computed (denoted as \(\mathbf{W_{proj}^{t}}\)), we update the knowledge parameters in the queue according to the following steps to capture the semantic correlation of sequential edits and alleviate degradation in the general capabilities of LLMs.

\noindent\textbf{Step 1: Enqueuing Located Parameters.}  
We append the current edited knowledge parameter \(\mathbf{W_{proj}^{t}}\) to the end of the weight queue \(Q = \langle W_{proj}^{1}, W_{proj}^{2}, \dots, W_{proj}^{m} \rangle \quad (m \geq 0)\),  
where \(m\) denotes the number of located parameters currently in the queue \(Q\).  
Thus, the updated queue order is  
$Q = \langle W_{proj}^{1}, W_{proj}^{2}, \dots, W_{proj}^{m}, \mathbf{W_{proj}^{t}} \rangle$.

\noindent\textbf{Step 2: Selecting Top-K Parameters.}  
Following the FIFO principle, the parameter bias at the front of the queue caused by the semantic gap with \(\mathbf{W_{proj}^{t}}\) has a greater impact on editing effectiveness~\cite{DBLP:conf/acl/GuptaRA24,DBLP:journals/corr/abs-2403-07175}.  
Considering the trade-off between queue length and memory storage, we first identify the \texttt{Top-K} elements in the queue that need to be aligned.  
Specifically, we calculate the similarity in the editing parameter space between the current sample \(t\) and others using the Euclidean distance:
\begin{equation}
\label{eq_13}
    d_i = \| \mathbf{W_{proj}^t} - W_{proj}^i \|_2, \quad i = 1, 2, \dots, m
\end{equation}
\begin{equation}
    \mathcal{I}_{\text{topK}} = \{ \sigma(1), \sigma(2), \dots, \sigma(K) \}, \quad K \leq m.
\end{equation}
The function \(\sigma(\cdot)\) denotes the sorting order such that 
$d_{\sigma(1)} \leq d_{\sigma(2)} \leq \cdots \leq d_{\sigma(m)}$
and \(\mathcal{I}_{\text{topK}}\) is the index set of the \texttt{Top-K} parameters.  
Note that only when the similarity \(d_i\) is below a threshold \(\eta_{que}\) do we consider there to be a strong semantic correlation between the two parameter matrices, and thus align the semantic spaces between \(\mathbf{W_{proj}^t}\) and \(W_{proj}^i\).

\noindent\textbf{Step 3: Editing with Self-Correction.}  
To iteratively update the dependency of previous data on the current editing sample in the SME task, we observe from Figure~\ref{motivation_res} that the parameters needing updates typically arise by linking the object \(o_*^t\) of the newly edited knowledge to the relation \(r^i\) of the previously edited knowledge. In general, it suffices to replace the object of the \(t\)-th sample with the subject of the \(i\)-th sample. Other parameters in the LLM and the queue \(Q\) remain frozen to preserve the general capabilities of the LLM. The effect of semantic superposition is analogous to prior work on embedding translation operations~\cite{DBLP:conf/nips/BordesUGWY13,DBLP:conf/aaai/LinLSLZ15}.

Concretely, we leverage the \texttt{Top-K} located parameters \(\{W_{proj}^{1}, W_{proj}^{2}, \ldots, W_{proj}^{K}\}\) and align them semantically with the current edited knowledge \(\mathbf{W_{proj}^{t}}\).  
The update rule for previously semantically dependent editing samples is:
\begin{gather}
    \Delta_W = v_*^t \oplus h_r^{i}, \quad i \in \mathcal{I}_{\text{topK}} \\
    W_{proj}^{i'} = \sigma\left(W_{proj}^{i} \parallel \Delta_W + b' \right)
\end{gather}
where \(\oplus\) denotes element-wise addition and \(\parallel\) denotes concatenation. \(W_{proj}^{i'}\) is the self-corrected editing parameter in the queue that will be aligned back into the LLM. Due to the limited length of the queue, the oldest knowledge parameter is dequeued in the next step after each self-correction.

\noindent\textbf{Step 4: Dequeuing Located Parameters.}  
According to the FIFO principle, the edited triple at the head of the queue, \(W_{proj}^{head}\), is the farthest from the current editing sample \(W_{proj}^t\) and thus has the smallest semantic correlation. We use Eq.~\ref{eq_13} to calculate the similarity \(d_{head}\) between them to determine whether to dequeue:
\begin{equation}
    \text{Dequeue}(Q) \triangleq \begin{cases}
    \langle Q \backslash \{ W_{proj}^{head} \} \rangle & d_{head} < \eta_{deq}, \\
    Q & \text{otherwise}.
    \end{cases}
\end{equation}
where \(\eta_{deq}\) is the dequeue threshold hyperparameter.

\subsection{Model Training}  
To satisfy the three desired properties for SME (reliability, generality, and locality), we adopt the same loss formulations with respect to \(T\) SME facts as in previous model editing works~\cite{DBLP:conf/iclr/MengSABB23, DBLP:conf/acl/ZhangCL0HHXH24}.  
The total loss is the sum of the structural editing loss and the typical model editing losses used in prior studies, expressed as:  
$\mathcal{L}_{total} = \alpha_1 \cdot \mathcal{L}_{ed}(f_T) + \alpha_2 \cdot \mathcal{L}_{st}(f_T)$,
where \(\alpha_i\) are coefficients satisfying \(\sum_{i=1}^2 \alpha_i = 1\).

\section{Experiments}
Due to space limitations, the descriptions of datasets, baselines, and implementations are provided in Appendix~\ref{implementation_details}.

\begin{table*}[!tb]
\footnotesize
\centering
\setlength{\tabcolsep}{4.3pt}
\renewcommand{\arraystretch}{0.85}
\begin{tabular}{cccccccccccccc}
\midrule
\multirow{2}{*}{\textbf{Backbone}}  & \multirow{2}{*}{\textbf{Editor}} & \multicolumn{4}{c}{\textbf{ZSRE}}                 & \multicolumn{4}{c}{\textbf{CounterFact}}                   & \multicolumn{4}{c}{\textbf{RIPE}}                 \\
                                &                                       & \textbf{Rel.} & \textbf{Gen.} & \textbf{Loc.} & \textbf{Avg.} & \textbf{Rel.} & \textbf{Gen.} & \textbf{Loc.} & \textbf{Avg.} & \textbf{Rel.} & \textbf{Gen.} & \textbf{Loc.} & \textbf{Avg.} \\ \midrule

\multirow{14}{*}{\makecell[c]{GPT-J \\ (6B)} }

&FT&4.3&3.0&0.1&2.5$_{(\pm0.1)}$&12.9&5.1&1.1&6.4$_{(\pm0.1)}$&3.1&0.9&0.8&1.6$_{(\pm0.0)}$\\ \cmidrule{2-14}
&KN&0.8&0.0&2.2&1.0$_{(\pm0.0)}$&0.1&0.4&1.0&0.5$_{(\pm0.0)}$&0.0&0.0&0.0&0.0$_{(\pm0.0)}$\\
&ROME&57.2&53.9&29.9&47.0$_{(\pm1.1)}$&0.2&0.2&0.0&0.1$_{(\pm0.0)}$&47.5&16.9&13.4&26.0$_{(\pm0.5)}$\\
&MEMIT&56.8&54.6&54.9&55.4$_{(\pm1.3)}$&{42.3}&36.4&30.7&49.8$_{(\pm1.3)}$&0.0&0.0&0.0&0.0$_{(\pm0.0)}$\\
&$\text{KE}$&0.0&0.0&1.1&0.4$_{(\pm0.0)}$&0.0&0.0&0.1&0.0$_{(\pm0.0)}$&0.0&0.0&0.2&0.1$_{(\pm0.0)}$\\
&$\text{MEND}$&0.0&0.0&0.0&0.0$_{(\pm0.0)}$&0.0&0.0&0.0&0.0$_{(\pm0.0)}$&0.2&0.1&0.1&0.1$_{(\pm0.0)}$\\
&$\text{MALMEN}$&43.0&35.1&39.3&39.1$_{(\pm0.4)}$&15.0&12.4&25.1&17.5$_{(\pm0.4)}$&31.1&19.1&35.3&28.5$_{(\pm0.6)}$\\
&$\text{DAFNet}$&{60.0}&{57.6}&{88.0}&{68.5}$_{(\pm1.9)}$&53.1&{38.1}&{82.3}&{57.8}$_{(\pm1.2)}$&{48.3}&{31.3}&{57.3}&{45.6}$_{(\pm1.2)}$\\
&$\text{AlphaEdit}$&{40.2}&{35.2}&{62.1}&{45.8}$_{(\pm1.3)}$&37.5&{21.4}&{59.5}&{39.5}$_{(\pm1.5)}$&{31.6}&{19.8}&{43.2}&{31.5}$_{(\pm1.1)}$\\ \cmidrule{2-14}
&TP&45.7&40.4&10.5&32.2$_{(\pm0.8)}$&47.3&17.0&1.4&21.9$_{(\pm0.7)}$&48.1&29.1&15.2&30.8$_{(\pm0.6)}$\\
&GRACE&56.2&51.3&28.4&45.3$_{(\pm1.2)}$&0.3&0.4&0.1&0.3$_{(\pm0.1)}$&46.7&16.3&13.8&25.6$_{(\pm0.7)}$\\ \cmidrule{2-14}
&$\text{MELO}$&48.7&42.6&68.8&53.4$_{(\pm1.3)}$&41.6&28.3&65.1&45.0$_{(\pm0.5)}$&34.2&23.6&48.3&35.4$_{(\pm1.1)}$\\
&$\text{LTE}$&53.2&46.5&73.5&57.7$_{(\pm1.4)}$&46.6&32.5&73.1&50.7$_{(\pm1.0)}$&41.2&28.7&53.0&41.0$_{(\pm0.6)}$\\
& \cellcolor[HTML]{C0C0C0} $\text{QueueEDIT}$ \cellcolor[HTML]{C0C0C0} &\cellcolor[HTML]{C0C0C0} \textbf{64.5}& \cellcolor[HTML]{C0C0C0} \textbf{60.1}&\cellcolor[HTML]{C0C0C0} \textbf{92.6}&\cellcolor[HTML]{C0C0C0} \textbf{72.4}$_{(\pm1.1)}$&\cellcolor[HTML]{C0C0C0} \textbf{64.8}&\cellcolor[HTML]{C0C0C0} \textbf{43.2}&\cellcolor[HTML]{C0C0C0} \textbf{88.4}&\cellcolor[HTML]{C0C0C0} \textbf{65.5}$_{(\pm1.1)}$&\cellcolor[HTML]{C0C0C0} \textbf{51.7}&\cellcolor[HTML]{C0C0C0} \textbf{37.9}&\cellcolor[HTML]{C0C0C0} \textbf{62.2}&\cellcolor[HTML]{C0C0C0} \textbf{50.6}$_{(\pm0.7)}$\\ \midrule

\multirow{14}{*}{\makecell[c]{LLAMA3 \\ (8B)}}

&FT&8.4&7.2&5.1&6.9$_{(\pm0.1)}$&2.0&0.6&2.2&1.6$_{(\pm0.0)}$&3.2&1.5&2.7&2.5$_{(\pm0.0)}$\\ \cmidrule{2-14}
&KN&0.5&0.8&0.5&0.6$_{(\pm0.1)}$&0.9&0.2&0.1&0.4$_{(\pm0.1)}$&0.6&0.4&0.5&0.5$_{(\pm0.1)}$\\
&ROME&2.1&2.0&1.1&1.7$_{(\pm0.1)}$&0.7&0.6&0.6&0.6$_{(\pm0.1)}$&0.5&0.5&0.5&0.5$_{(\pm0.1)}$\\
&MEMIT&0.7&0.7&0.6&0.7$_{(\pm0.1)}$&0.6&0.6&1.5&0.9$_{(\pm0.1)}$&0.1&0.5&0.6&0.4$_{(\pm0.1)}$\\

&$\text{KE}$&0.3&0.8&0.6&0.5$_{(\pm0.1)}$&0.5&0.5&0.8&0.6$_{(\pm0.1)}$&0.0&0.2&0.1&0.1$_{(\pm0.1)}$\\
&$\text{MEND}$&0.8&0.1&0.3&0.4$_{(\pm0.1)}$&0.3&0.8&0.4&0.5$_{(\pm0.1)}$&0.0&0.2&0.6&0.3$_{(\pm0.1)}$\\
&$\text{MALMEN}$&32.9&29.4&29.0&30.4$_{(\pm0.6)}$&16.7&17.3&23.4&19.1$_{(\pm0.3)}$&43.2&39.3&39.4&40.6$_{(\pm0.9)}$\\
&$\text{DAFNet}$&51.4&49.5&94.5&65.1$_{(\pm1.3)}$&51.3&36.7&77.8&55.3$_{(\pm1.6)}$&45.0&35.2&86.7&55.6$_{(\pm0.9)}$\\
&$\text{AlphaEdit}$&34.2&28.6&71.8&44.9$_{(\pm1.1)}$&34.6&23.1&65.0&40.9$_{(\pm1.3)}$&29.1&22.4&71.1&40.9$_{(\pm1.1)}$\\ \cmidrule{2-14}
&TP&48.2&45.0&5.1&32.8$_{(\pm0.6)}$&65.6&33.4&12.3&37.1$_{(\pm0.9)}$&43.2&27.7&10.8&27.2$_{(\pm0.6)}$\\
&GRACE&2.0&2.1&1.3&1.8$_{(\pm0.1)}$&0.6&0.7&0.6&0.6$_{(\pm0.1)}$&0.3&0.2&0.5&0.3$_{(\pm0.1)}$\\ \cmidrule{2-14}
&$\text{MELO}$&39.2&32.5&75.3&49.0$_{(\pm1.4)}$&40.3&27.1&70.2&45.9$_{(\pm1.3)}$&34.5&26.2&75.8&45.5$_{(\pm1.6)}$\\
&$\text{LTE}$&43.7&40.6&82.3&55.5$_{(\pm1.0)}$&44.5&31.7&73.5&49.9$_{(\pm1.4)}$&42.0&32.6&81.5&52.0$_{(\pm1.2)}$\\ 
&\cellcolor[HTML]{C0C0C0} $\text{QueueEDIT}$&\cellcolor[HTML]{C0C0C0} \textbf{56.1}&\cellcolor[HTML]{C0C0C0} \textbf{55.0}&\cellcolor[HTML]{C0C0C0} \textbf{96.2}&\cellcolor[HTML]{C0C0C0} \textbf{69.1}$_{(\pm1.3)}$&\cellcolor[HTML]{C0C0C0} \textbf{60.5}&\cellcolor[HTML]{C0C0C0} \textbf{44.7}&\cellcolor[HTML]{C0C0C0} \textbf{80.1}&\cellcolor[HTML]{C0C0C0} \textbf{61.8}$_{(\pm0.7)}$&\cellcolor[HTML]{C0C0C0} \textbf{49.1}&\cellcolor[HTML]{C0C0C0} \textbf{44.4}&\cellcolor[HTML]{C0C0C0} \textbf{90.0}&\cellcolor[HTML]{C0C0C0} \textbf{61.2}$_{(\pm0.8)}$\\
\midrule
\end{tabular}
\caption{Overall results of \texttt{QueueEDIT} under 1000 edits. ``Rel.'', ``Gen.'', and ``Loc.'' represent the editing metrics, respectively.  
Due to space limitations, the results for 1, 10, and 100 edits are provided in Appendix~\ref{apendix_seq_res}.}
\label{main_exp}
\vspace{-0.5cm}
\end{table*}

\subsection{Main Results}

\noindent\textbf{Model Editing Results.}  
We evaluate \texttt{QueueEDIT} on three benchmarks and compare performance with 1000 edits\footnote{The results for 1, 10, and 100 edits are shown in Appendix~\ref{apendix_seq_res}.}.  
Table~\ref{main_exp} shows the overall performance. We observe the following:  
(1) Methods based on modifying parameters show a sharp decrease in performance as the number of edits increases. We hypothesize that meta-learning methods do not account for the sequential modeling of facts. The weight update training of located neurons in meta networks is carried out separately with a small amount of data, so these methods cannot capture the relationships between sequential editing data. Although locate-then-edit models have less performance degradation with increasing edits compared to meta-learning-based models, we find they are more sensitive to the backbone. When using LLAMA3 as the backbone, the accuracy of such models is nearly zero, especially in long editing scenarios (e.g., 1000 edits). KN~\cite{DBLP:conf/acl/DaiDHSCW22} achieves editing by performing excessive activation when dealing with multiple facts.  
(2) Methods that add extra parameters do not modify the original parameters inside the model, but add new neurons or modules at the located positions, which can partially capture the inherent connections among sequences. However, as the number of edits increases, the model’s performance gradually decreases regardless of the backbone.  
(3) Retrieve-data methods freeze the original parameters of LLMs and do not add additional parameters; retrieving external data only for the current editing sample cannot effectively correlate or model previous edits.  
(4) \texttt{QueueEDIT} modifies only parameters for the current edit, while using a queue structure to self-correct correlations between previously edited data. Therefore, the model not only achieves strong editing results but also maintains good stability during long sequence editing.

\begin{figure}[!t]
  \centering
  \includegraphics[width=9cm]{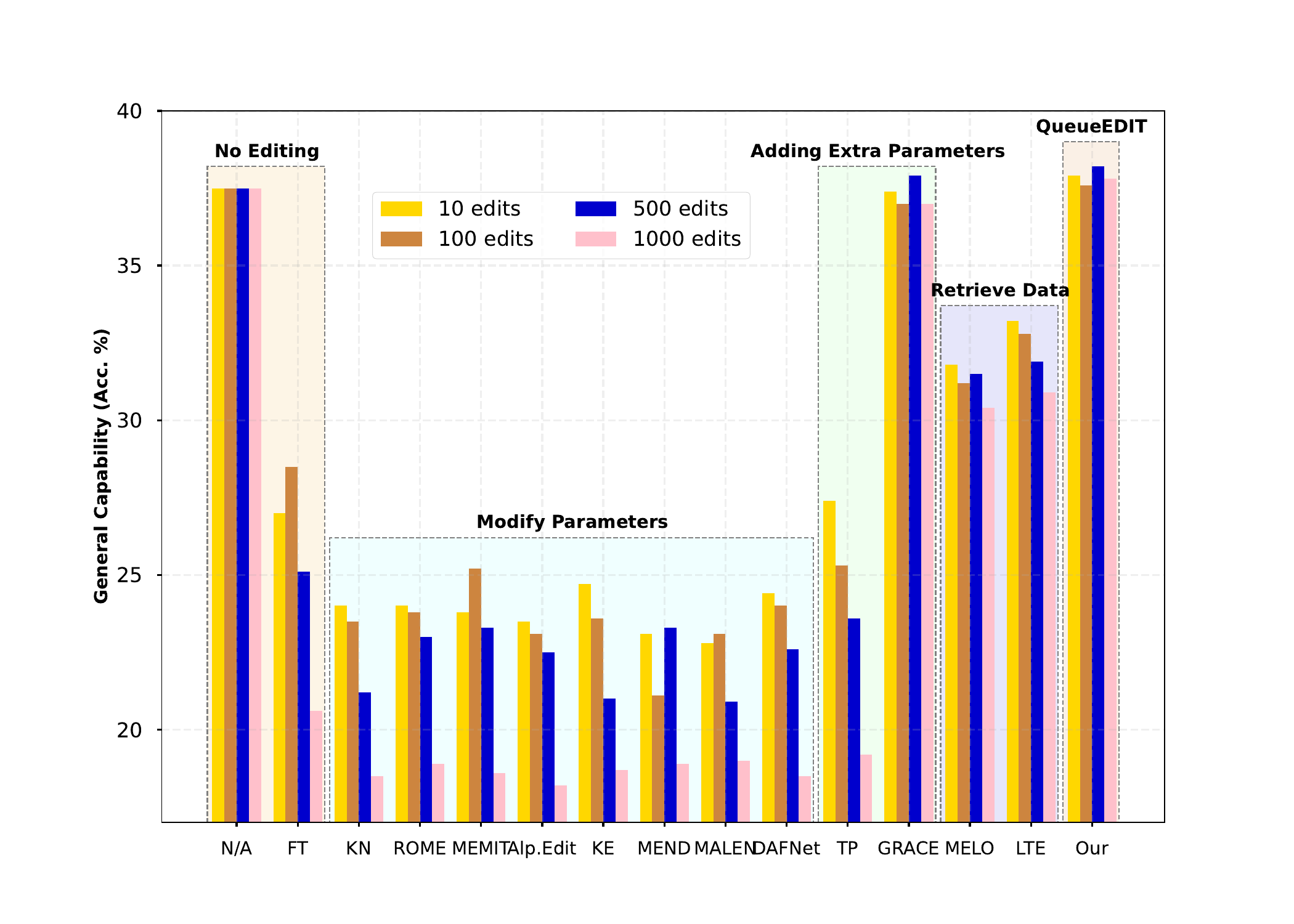}
  \caption{The ``improvement/degeneration'' of general capability after performing the SME task across various editing paradigms. ``N/A'' indicates the original results of the LLMs.}
  \label{general_benchmark_res}
  \vspace{-0.7cm}
\end{figure}

\noindent\textbf{General Capabilities of LLMs.}  
We further investigate the extent to which these editors influence the general capabilities of LLMs during SME. Figure~\ref{general_benchmark_res} shows the averaged accuracy results for LLaMA3 (8B) after \(10\), \(100\), \(500\), and \(1000\) edits over public LLM benchmarks.  
The benchmark datasets typically consist of QA tasks similar to the editing samples, thereby alleviating bias due to differences in data format between training and testing.  
It is observed that previous SME methods harm the general capabilities of LLMs as the number of edits increases. Grace~\cite{GRACE} leverages extra datasets and adapters to preserve general performance. Retrieval-based methods perform relatively better than other baselines.  
We conjecture that the retrieved external data may be mixed with diverse types of knowledge, which helps alleviate the loss of model generality.  
In contrast, \texttt{QueueEDIT} performs self-correction only on the parameters corresponding to previously associated edited data during the queue update process, while freezing other parameters in both the queue and LLMs.  
Therefore, our model better generalizes on other open-domain questions.

\begin{table*}[!tb]
\scriptsize
\centering
\setlength{\tabcolsep}{4.3pt}
\begin{tabular}{ccccccccccccccc}
\toprule
\multirow{2}{*}{\textbf{Backbone}}  & \multirow{2}{*}{\textbf{QL}} & \multirow{2}{*}{\textbf{Mem. (GB)}} & \multicolumn{4}{c}{\textbf{ZSRE}}                 & \multicolumn{4}{c}{\textbf{CounterFact}}                   & \multicolumn{4}{c}{\textbf{RIPE}}                 \\
                                &                              &                                  & \textbf{Rel.} & \textbf{Gen.} & \textbf{Loc.} & \textbf{Avg.} & \textbf{Rel.} & \textbf{Gen.} & \textbf{Loc.} & \textbf{Avg.} & \textbf{Rel.} & \textbf{Gen.} & \textbf{Loc.} & \textbf{Avg.} \\ \midrule

\multirow{6}{*}{\makecell[c]{LLAMA3 \\ (8B)}} & 10\%&12.5&49.4&50.4&92.0&63.9$_{(\pm0.5)}$&47.5&38.8&70.3&52.2$_{(\pm0.3)}$&45.0&41.0&80.1&55.4$_{(\pm0.5)}$\\
&30\%&21.8&\cellcolor[HTML]{C0C0C0} \textbf{56.1}&\cellcolor[HTML]{C0C0C0} \textbf{55.0}&\cellcolor[HTML]{C0C0C0} \textbf{96.2}&\cellcolor[HTML]{C0C0C0} \textbf{69.1}$_{(\pm1.3)}$&\cellcolor[HTML]{C0C0C0} \textbf{60.5}&\cellcolor[HTML]{C0C0C0} \textbf{44.7}&\cellcolor[HTML]{C0C0C0} \textbf{80.1}&\cellcolor[HTML]{C0C0C0} \textbf{61.8}$_{(\pm0.7)}$&\cellcolor[HTML]{C0C0C0} \textbf{49.1}&\cellcolor[HTML]{C0C0C0} \textbf{44.4}&\cellcolor[HTML]{C0C0C0} \textbf{90.0}&\cellcolor[HTML]{C0C0C0} \textbf{61.2}$_{(\pm0.8)}$\\
&50\%&32.3&49.0&51.8&92.5&64.4$_{(\pm0.5)}$&52.4&38.2&73.2&54.6$_{(\pm0.4)}$&45.6&39.7&83.5&56.3$_{(\pm0.2)}$\\ \cmidrule{2-15}
& \multicolumn{14}{c}{ \cellcolor[HTML]{C0C0C0} Memory Consumption of Baselines} \\
&$\text{FT}$&$\text{KN}$&$\text{ROME}$&$\text{MEMIT}$&$\text{KE}$&$\text{MEND}$&$\text{MALMEN}$&$\text{DAFNet}$&$\text{AlphaEdit}$&$\text{TP}$&$\text{Grace}$&$\text{MELO}$&$\text{LTE}$&-\\
&60.2&31.4&18.3&25.6&13.9&16.2&13.7&30.5&26.7&14.8&47.1&36.5&28.3&-\\
\bottomrule                     
\end{tabular}
\caption{Results of \texttt{QueueEDIT} with different queue lengths in 1000 edits. ``QL'' indicates the queue length.}
\label{memory_length}
\vspace{-0.5cm}
\end{table*}

\begin{table}[]
\footnotesize
\begin{tabular}{lccc|cccc}
\toprule
\multirow{2}{*}{\textbf{Modules}} & \multicolumn{3}{c|}{\textbf{100 edits}}                      & \multicolumn{3}{c}{\textbf{1000 edits}}                     & \multirow{2}{*}{\textbf{Avg.}} \\ \cmidrule{2-7}
                     & \multicolumn{1}{l}{\textbf{Rel.}} & \multicolumn{1}{l}{\textbf{Gen.}} & \multicolumn{1}{l|}{\textbf{Loc.}} & \multicolumn{1}{l}{\textbf{Rel.}} & \multicolumn{1}{l}{\textbf{Gen.}} & \multicolumn{1}{l}{\textbf{Loc.}}  &                      \\ \midrule
                  \rowcolor[HTML]{C0C0C0}  Ours                             & 77.1 & 56.5 & 91.1    & 55.2 & 54.7 & 88.8 & 70.6   \\ \midrule
                     w/o $\mathcal{L}_{st}$   &  72.4       & 51.8         & 84.9 & 48.3         & 49.9         & 82.3 & 64.9   \\
                     w/o Queue   & 59.0         & 45.2         & 73.8    & 40.4        & 39.5          & 69.7   & 54.6     \\
                     w/o Top-K   &  73.5         & 53.6         & 88.0   & 51.3        & 50.5         & 86.1   & 67.2     \\
                     \bottomrule
\end{tabular}
\caption{Ablation study of \texttt{QueueEDIT}.}
\label{tab_ablation}
\vspace{-0.5cm}
\end{table}

\subsection{Ablation Study}
We conduct an ablation study using LLaMA3 (8B), with averaged results presented in Table~\ref{tab_ablation}.  
The structural mapping editing is replaced with the original editing representation \(k_*\), which mixes the subject and relation~\cite{DBLP:conf/nips/MengBAB22}.  
When we remove the queue-based self-correction, the overall system degrades to a basic locate-the-editing paradigm equipped with the structural editing loss.  
In addition, we evaluate the effect of selecting parameters by removing the \texttt{Top-K} calculation and instead using random selection.

We observe that without queue-based learning, performance decreases significantly due to the lack of alignment among sequential editing samples in the editing parameters.  
The decrease observed when the structural mapping loss is removed indicates that editing knowledge through structural semantics is more effective than directly mixing triplet data together, as in previous works~\cite{DBLP:conf/nips/MengBAB22,DBLP:conf/iclr/MengSABB23}.  
After removing the \texttt{Top-K} selection mechanism, the decrease in model performance suggests that randomly selecting semantically unrelated triple parameters for alignment disrupts the consistency of parameter learning and leads to reduced model editing capability.

\begin{figure}[!t]
  \centering
  \includegraphics[width=8cm]{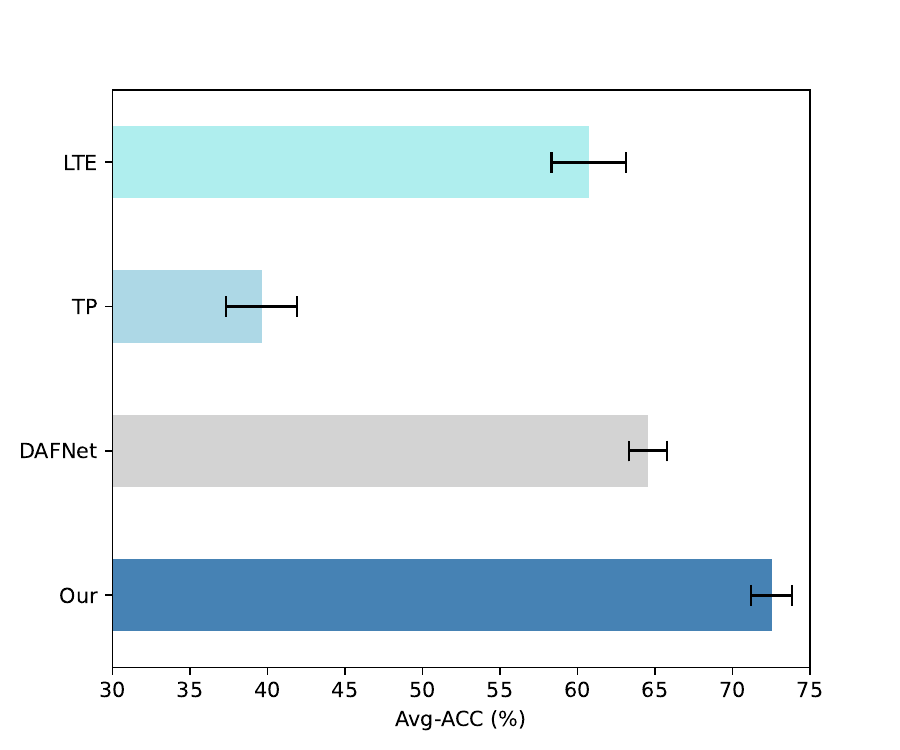}
  \vspace{-0.2cm}
  \caption{Results using Qwen2.5-14B as the backbone.}
  \label{larger_model}
\vspace{-0.5cm}
\end{figure}

\subsection{Detailed Analysis}

\noindent\textbf{Queue Length.}  
We analyze memory usage and performance changes as the queue length increases. Other methods, especially those with additional parameters, require extra memory to train the knowledge adapter module. We set the queue length to 10\%, 30\%, and 50\% of 1000 edits to evaluate editing performance and memory consumption.  
For memory consumption, we compare the number of parameters used by the additional editor regardless of the frozen LLMs.  
As shown in Table~\ref{memory_length}, although \texttt{QueueEDIT} consumes slightly more memory compared to baseline methods, its performance improves significantly as the number of edits increases.  
Additionally, we find that longer queue lengths lead to less consistent editing performance, possibly due to the introduction of more irrelevant knowledge parameters and semantic noise.

\noindent\textbf{Qualitative Analysis of Queue-Based Self-Correction.}  
We present case studies demonstrating the queue-based self-correction of relevant knowledge within \texttt{QueueEDIT}. Due to space limitations, please refer to Appendix~\ref{case_study} and~\ref{queue_self_corr}.

\noindent\textbf{Larger LLMs as Backbones.}  
The mainstream LLM backbones used in model editing are GPT-J (6B) and LLaMA3 (8B)~\cite{DBLP:conf/icml/MitchellLBMF22,DBLP:conf/emnlp/MadaanTCY22,DBLP:journals/corr/abs-2305-12740,DBLP:conf/acl/ZhangCL0HHXH24,DBLP:conf/iclr/FangJWMSW0C25}.  
We further test our model’s effectiveness using a larger backbone. Due to limited GPU resources, we select the larger-scale Qwen2.5-14B model~\cite{qwen25}, a state-of-the-art LLM, for our experiments.  
The experimental setup includes several competitive editing paradigms (i.e., DAFNet, TP, and LTE) with 1000 sequential edits. We report average results on three editing properties using F1 (\%).

From Figure~\ref{larger_model}, we observe that:  
(1) Utilizing a larger-scale backbone model for SME further enhances editing performance. This improvement is likely due to the increased capacity for internal parameterized knowledge storage in larger models compared to smaller LLMs, allowing more effective collaborative modeling via both internal and external knowledge sources. 
(2) With the advantage of a larger backbone, our \texttt{QueueEDIT} model demonstrates significant performance improvements over baselines.

\vspace{-0.2cm}

\section{Conclusion}

In this paper, we propose \texttt{QueueEDIT}, a novel queue-based self-correction framework designed for SME in LLMs. Our approach introduces a structural mapping editing loss to effectively associate knowledge triplets with knowledge-sensitive neurons within Transformer layers. By storing and dynamically aligning editing parameters in a queue, \texttt{QueueEDIT} captures long-sequence dependencies and selectively updates relevant parameters, thereby mitigating the negative impact of parameter bias on the general capabilities of LLMs. Extensive experiments demonstrate that \texttt{QueueEDIT} significantly outperforms strong baselines across diverse SME scenarios, and effectively preserves the general language understanding abilities of LLMs throughout the editing process.

\bibliography{aaai2026}

\appendix

\begin{table*}[!tb]
\scriptsize
\centering
\setlength{\tabcolsep}{4.3pt}
\renewcommand{\arraystretch}{0.85}
\begin{tabular}{ccccccccccccccc}
\toprule
\multirow{2}{*}{\textbf{Backbone}} & \multirow{2}{*}{\textbf{\# Editing}} & \multirow{2}{*}{\textbf{Editor}} & \multicolumn{4}{c}{\textbf{ZSRE}}                 & \multicolumn{4}{c}{\textbf{CounterFact}}                   & \multicolumn{4}{c}{\textbf{RIPE}}                 \\
                                &                               &                                  & \textbf{Rel.} & \textbf{Gen.} & \textbf{Loc.} & \textbf{Avg.} & \textbf{Rel.} & \textbf{Gen.} & \textbf{Loc.} & \textbf{Avg.} & \textbf{Rel.} & \textbf{Gen.} & \textbf{Loc.} & \textbf{Avg.} \\ \midrule
\multirow{14}{*}{\makecell[c]{LLAMA3 \\ (8B)}}

&\multirow{14}{*}{1}&FT&53.3&53.7&92.6&66.5$_{(\pm0.3)}$&34.6&25.9&50.3&36.9$_{(\pm0.5)}$&49.7&34.2&71.5&51.8$_{(\pm0.5)}$\\
&&KN&20.7&21.3&52.9&31.6$_{(\pm0.1)}$&12.8&9.7&68.4&30.3$_{(\pm0.4)}$&22.3&15.9&55.9&31.4$_{(\pm0.5)}$\\
&&ROME&54.0&52.1&94.5&66.9$_{(\pm0.7)}$&41.6&22.3&92.3&52.1$_{(\pm0.8)}$&48.8&27.6&43.0&39.8$_{(\pm0.9)}$\\
&&MEMIT&50.2&49.9&92.4&64.2$_{(\pm0.5)}$&45.9&29.8&93.4&56.4$_{(\pm0.4)}$&58.9&30.1&39.2&42.7$_{(\pm0.3)}$\\
&&$\text{KE}$&13.4&9.1&91.1&37.9$_{(\pm0.2)}$&8.5&3.4&90.8&34.2$_{(\pm0.4)}$&10.4&4.9&43.3&19.5$_{(\pm0.2)}$\\
&&$\text{MEND}$&74.3&70.8&66.6&70.6$_{(\pm0.5)}$&81.6&67.7&77.6&75.6$_{(\pm0.3)}$&66.9&29.9&30.2&42.3$_{(\pm0.5)}$\\
&&$\text{MALMEN}$&66.9&68.3&44.2&59.8$_{(\pm0.5)}$&52.9&42.8&37.1&44.3$_{(\pm0.6)}$&52.0&34.3&21.0&35.8$_{(\pm0.9)}$\\
&&$\text{DAFNet}$&97.9&97.8&95.4&97.0$_{(\pm1.5)}$&92.5&87.2&94.8&91.5$_{(\pm0.8)}$&98.2&66.9&72.7&79.3$_{(\pm0.7)}$\\
&&$\text{AlphaEdit}$& 98.6&98.6&95.8&97.7$_{(\pm1.2)}$&93.3&\textbf{88.6}&95.7&92.5$_{(\pm0.9)}$&97.9&67.6&73.5&79.7$_{(\pm0.5)}$\\ \cmidrule{3-15}
&&TP&86.9&84.5&86.9&86.1$_{(\pm1.4)}$&91.9&69.1&39.5&66.8$_{(\pm0.3)}$&77.5&55.6&51.8&61.6$_{(\pm0.5)}$\\
&&GRACE&52.8&51.3&96.2&66.8$_{(\pm0.9)}$&45.1&29.0&93.9&56.0$_{(\pm0.5)}$&57.2&31.1&41.6&43.3$_{(\pm0.2)}$\\ \cmidrule{3-15}
&&$\text{MELO}$&97.3&96.5&92.2&95.3$_{(\pm1.1)}$&91.4&84.9&92.6&89.6$_{(\pm1.0)}$&93.3&64.0&68.8&75.4$_{(\pm0.5)}$\\
&&$\text{LTE}$&97.7&96.9&93.5&96.0$_{(\pm1.1)}$&92.2&85.8&93.4&90.5$_{(\pm0.7)}$&94.6&65.4&70.1&76.7$_{(\pm1.3)}$\\
&& \cellcolor[HTML]{C0C0C0} $\text{QueueEDIT}$ \cellcolor[HTML]{C0C0C0} &\cellcolor[HTML]{C0C0C0} \textbf{98.6}&\cellcolor[HTML]{C0C0C0} \textbf{99.6}&\cellcolor[HTML]{C0C0C0} \textbf{97.0}&\cellcolor[HTML]{C0C0C0} \textbf{98.4}$_{(\pm0.5)}$&\cellcolor[HTML]{C0C0C0} \textbf{93.9}&\cellcolor[HTML]{C0C0C0} 88.4&\cellcolor[HTML]{C0C0C0} \textbf{96.1}&\cellcolor[HTML]{C0C0C0} \textbf{92.8}$_{(\pm0.6)}$&\cellcolor[HTML]{C0C0C0} \textbf{98.2}&\cellcolor[HTML]{C0C0C0} \textbf{68.8}&\cellcolor[HTML]{C0C0C0} \textbf{76.2}&\cellcolor[HTML]{C0C0C0} \textbf{81.1}$_{(\pm0.9)}$\\ \bottomrule
                                
\end{tabular}
\caption{The overall results in single-turn editing.}
\label{single_turn_main_exp}
\vspace{-0.5cm}

\end{table*}

\begin{table*}[!tb]
\scriptsize
\centering
\setlength{\tabcolsep}{4.3pt}
\begin{tabular}{ccccccccccccccc}
\midrule
\multirow{2}{*}{\textbf{Backbone}} & \multirow{2}{*}{\textbf{\# Editing}} & \multirow{2}{*}{\textbf{Editor}} & \multicolumn{4}{c}{\textbf{ZSRE}}                 & \multicolumn{4}{c}{\textbf{CounterFact}}                   & \multicolumn{4}{c}{\textbf{RIPE}}                 \\
                                &                               &                                  & \textbf{Rel.} & \textbf{Gen.} & \textbf{Loc.} & \textbf{Avg.} & \textbf{Rel.} & \textbf{Gen.} & \textbf{Loc.} & \textbf{Avg.} & \textbf{Rel.} & \textbf{Gen.} & \textbf{Loc.} & \textbf{Avg.} \\ \midrule

\multirow{28}{*}{\makecell[c]{GPT-J \\ (6B)} }
&\multirow{14}{*}{10}&FT&10.3&10.8&0.3&7.1$_{(\pm0.1)}$&56.2&24.2&2.1&27.5$_{(\pm0.5)}$&7.8&4.3&1.4&4.5$_{(\pm0.1)}$\\
&&KN&1.0&1.1&1.9&1.3$_{(\pm0.0)}$&1.2&0.7&2.3&1.4$_{(\pm0.0)}$&0.1&0.3&0.2&0.2$_{(\pm0.0)}$\\
&&ROME&81.1&78.8&94.6&84.8$_{(\pm1.7)}$&95.9&59.4&90.0&81.8$_{(\pm1.9)}$&98.2&41.9&39.1&59.7$_{(\pm0.8)}$\\
&&MEMIT&82.1&76.0&94.7&84.2$_{(\pm2.0)}$&96.0&38.1&\textbf{95.5}&76.5$_{(\pm2.5)}$&98.5&37.7&47.3&61.2$_{(\pm1.2)}$\\
&&$\text{KE}$&0.0&0.0&0.7&0.3$_{(\pm0.0)}$&0.0&0.0&0.2&0.1$_{(\pm0.0)}$&0.0&0.0&0.1&0.0$_{(\pm0.0)}$\\
&&$\text{MEND}$&0.4&0.4&0.5&0.4$_{(\pm0.0)}$&0.6&0.2&0.2&0.3$_{(\pm0.0)}$&0.0&0.0&0.0&0.0$_{(\pm0.0)}$\\
&&$\text{MALMEN}$&99.1&95.3&92.8&95.8$_{(\pm1.6)}$&90.0&32.9&77.1&66.7$_{(\pm2.2)}$&89.7&52.1&51.3&64.4$_{(\pm1.8)}$\\
&&$\text{DAFNET}$&99.6&97.6&94.8&97.3$_{(\pm1.5)}$&96.2&65.8&85.2&82.4$_{(\pm1.6)}$&98.7&57.6&57.8&71.4$_{(\pm1.6)}$\\
&&$\text{AlphaEdit}$&99.9&98.1&95.2&97.7$_{(\pm1.1)}$&96.8&66.5&86.8&83.4$_{(\pm1.0)}$&98.9&58.2&59.0&72.0$_{(\pm1.4)}$\\ \cmidrule{3-15}
&&TP&85.2&78.3&77.2&80.2$_{(\pm1.2)}$&96.0&54.3&3.6&51.3$_{(\pm1.2)}$&80.8&56.7&32.4&56.6$_{(\pm1.7)}$\\
&&GRACE&81.8&78.4&94.5&84.9$_{(\pm1.6)}$&95.2&60.3&91.2&82.2$_{(\pm1.6)}$&98.0&40.9&38.7&59.2$_{(\pm0.4)}$\\ \cmidrule{3-15}
&&$\text{MELO}$&98.2&96.5&92.8&95.8$_{(\pm0.9)}$&95.2&63.6&84.5&81.1$_{(\pm2.1)}$&96.6&51.9&55.7&68.1$_{(\pm0.8)}$\\
&&$\text{LTE}$&98.7&96.1&93.2&96.0$_{(\pm1.3)}$&94.3&60.6&82.3&79.1$_{(\pm1.0)}$&94.5&48.4&53.6&65.5$_{(\pm0.9)}$\\
&&\cellcolor[HTML]{C0C0C0}$\text{QueueEDIT}$\cellcolor[HTML]{C0C0C0}&\cellcolor[HTML]{C0C0C0}\textbf{99.9}&\cellcolor[HTML]{C0C0C0}\textbf{98.7}&\cellcolor[HTML]{C0C0C0}\textbf{95.9}&\cellcolor[HTML]{C0C0C0}\textbf{98.2}$_{(\pm0.6)}$&\cellcolor[HTML]{C0C0C0}\textbf{97.5}&\cellcolor[HTML]{C0C0C0}\textbf{68.2}&\cellcolor[HTML]{C0C0C0}89.4&\cellcolor[HTML]{C0C0C0}\textbf{85.0}$_{(\pm0.8)}$&\cellcolor[HTML]{C0C0C0}\textbf{99.8}&\cellcolor[HTML]{C0C0C0}\textbf{61.3}&\cellcolor[HTML]{C0C0C0}\textbf{61.4}&\cellcolor[HTML]{C0C0C0}\textbf{74.2}$_{(\pm1.1)}$\\ \cmidrule{2-15}

&\multirow{14}{*}{100}&FT&2.2&1.9&0.3&1.4$_{(\pm0.0)}$&35.9&10.8&1.6&16.1$_{(\pm0.3)}$&5.7&1.6&0.1&2.5$_{(\pm0.1)}$\\  \cmidrule{3-15}
&&KN&0.6&0.4&0.8&0.6$_{(\pm0.0)}$&0.2&0.5&0.8&0.5$_{(\pm0.0)}$&0.0&0.0&0.0&0.0$_{(\pm0.0)}$\\
&&ROME&77.4&75.6&85.0&79.3$_{(\pm2.2)}$&78.8&38.4&52.2&56.5$_{(\pm1.0)}$&\textbf{95.7}&36.0&32.2&54.6$_{(\pm1.0)}$\\
&&MEMIT&77.9&74.1&90.2&80.7$_{(\pm2.5)}$&\textbf{94.1}&40.2&85.1&73.1$_{(\pm1.3)}$&86.6&33.3&33.5&51.1$_{(\pm1.3)}$\\
&&$\text{KE}$&0.0&0.0&0.7&0.2$_{(\pm0.0)}$&0.0&0.0&0.1&0.0$_{(\pm0.0)}$&0.0&0.0&0.0&0.0$_{(\pm0.0)}$\\
&&$\text{MEND}$&0.2&0.1&0.0&0.1$_{(\pm0.0)}$&0.2&0.2&0.0&0.1$_{(\pm0.0)}$&0.0&0.0&0.1&0.0$_{(\pm0.0)}$\\
&&$\text{MALMEN}$&50.6&40.7&59.3&50.2$_{(\pm0.8)}$&29.7&31.8&68.0&43.2$_{(\pm0.4)}$&39.9&27.8&53.2&40.3$_{(\pm0.8)}$\\
&&$\text{DAFNET}$&89.5&76.5&90.2&85.4$_{(\pm1.6)}$&81.8&40.3&87.3&69.8$_{(\pm1.5)}$&78.5&38.9&64.4&60.6$_{(\pm1.5)}$\\
&&$\text{AlphaEdit}$&80.2&67.1&82.5&76.6$_{(\pm0.8)}$&70.7&35.9&74.8&60.5$_{(\pm1.2)}$&71.9&36.3&59.2&55.8$_{(\pm1.3)}$\\ \cmidrule{3-15}
&&TP&68.5&59.3&52.8&60.2$_{(\pm1.3)}$&76.0&31.9&2.2&36.7$_{(\pm0.8)}$&64.2&36.4&23.7&41.4$_{(\pm1.0)}$\\
&&GRACE&77.8&74.6&85.9&79.4$_{(\pm2.0)}$&76.3&39.2&51.6&55.7$_{(\pm0.8)}$&94.8&36.7&31.5&54.3$_{(\pm0.8)}$\\ \cmidrule{3-15}
&&$\text{MELO}$&78.4&65.3&82.9&75.5$_{(\pm0.8)}$&79.4&41.7&79.9&67.0$_{(\pm1.2)}$&74.6&30.1&52.3&52.3$_{(\pm0.7)}$\\
&&$\text{LTE}$&81.5&70.5&83.4&78.5$_{(\pm1.7)}$&78.3&38.7&80.6&65.9$_{(\pm1.0)}$&75.1&35.9&62.7&57.9$_{(\pm0.8)}$\\
& & \cellcolor[HTML]{C0C0C0} $ \text{QueueEDIT}$  & \cellcolor[HTML]{C0C0C0} \textbf{93.5}& \cellcolor[HTML]{C0C0C0}\textbf{78.6}& \cellcolor[HTML]{C0C0C0}\textbf{94.2}& \cellcolor[HTML]{C0C0C0}\textbf{88.8}$_{(\pm1.3)}$& \cellcolor[HTML]{C0C0C0} \textbf{92.6}& \cellcolor[HTML]{C0C0C0} \textbf{52.3}& \cellcolor[HTML]{C0C0C0} \textbf{90.1}& \cellcolor[HTML]{C0C0C0} \textbf{78.3}$_{(\pm0.9)}$&\cellcolor[HTML]{C0C0C0} 94.9&\cellcolor[HTML]{C0C0C0} \textbf{43.8}&\cellcolor[HTML]{C0C0C0} \textbf{68.7}&\cellcolor[HTML]{C0C0C0} \textbf{69.1}$_{(\pm0.9)}$\\

\midrule

\multirow{28}{*}{\makecell[c]{LLAMA3 \\ (8B)}}

&\multirow{14}{*}{10}&FT&38.7&38.1&58.5&45.1$_{(\pm0.9)}$&19.7&14.2&23.0&19.0$_{(\pm0.2)}$&31.0&22.5&28.7&27.4$_{(\pm0.7)}$\\
&&KN&0.9&0.9&1.4&1.1$_{(\pm0.1)}$&1.2&1.1&4.9&2.4$_{(\pm0.1)}$&0.9&0.9&0.8&0.9$_{(\pm0.1)}$\\
&&ROME&41.6&40.3&93.6&58.5$_{(\pm1.4)}$&39.1&25.6&84.3&49.7$_{(\pm1.0)}$&33.9&21.0&30.2&28.4$_{(\pm0.6)}$\\
&&MEMIT&24.8&24.7&51.8&33.8$_{(\pm0.9)}$&19.1&15.9&63.5&32.8$_{(\pm0.7)}$&18.9&14.2&10.8&14.6$_{(\pm0.3)}$\\
&&$\text{KE}$&1.2&1.2&1.9&1.4$_{(\pm0.1)}$&0.7&0.7&0.8&0.7$_{(\pm0.1)}$&0.8&0.9&1.7&1.1$_{(\pm0.1)}$\\
&&$\text{MEND}$&1.0&1.0&3.8&1.9$_{(\pm0.1)}$&0.7&0.7&0.8&0.7$_{(\pm0.1)}$&1.0&1.0&2.1&1.4$_{(\pm0.1)}$\\
&&$\text{MALMEN}$&96.7&89.0&93.3&93.0$_{(\pm1.8)}$&80.0&46.5&36.9&54.5$_{(\pm1.1)}$&85.2&48.2&71.6&68.3$_{(\pm2.3)}$\\
&&$\text{DAFNET}$&97.7&92.7&94.0&94.8$_{(\pm1.2)}$&87.8&60.3&86.5&78.2$_{(\pm1.4)}$&89.3&57.1&83.9&76.8$_{(\pm1.9)}$\\
&&$\text{AlphaEdit}$&98.2&92.3&94.5&95.0$_{(\pm1.1)}$&88.1&61.4&86.7&78.7$_{(\pm0.8)}$&90.1&58.6&84.5&77.7$_{(\pm1.2)}$\\ \cmidrule{3-15}
&&TP&57.8&53.1&37.4&49.4$_{(\pm1.1)}$&86.4&59.3&22.2&56.0$_{(\pm0.9)}$&63.9&42.0&31.1&45.7$_{(\pm0.8)}$\\
&&GRACE&43.0&40.4&93.1&58.8$_{(\pm1.5)}$&38.6&25.2&83.3&49.0$_{(\pm0.8)}$&31.9&21.5&29.8&27.7$_{(\pm0.5)}$\\ \cmidrule{3-15}
&&$\text{MELO}$&97.5&93.3&94.2&95.0$_{(\pm1.1)}$&86.2&57.8&83.5&75.8$_{(\pm1.7)}$&88.0&54.4&79.9&74.1$_{(\pm1.2)}$\\
&&$\text{LTE}$&96.9&93.5&93.2&94.5$_{(\pm1.6)}$&88.5&58.5&87.1&78.0$_{(\pm1.6)}$&87.2&55.7&80.2&74.4$_{(\pm1.1)}$\\
&&\cellcolor[HTML]{C0C0C0}$\text{QueueEDIT}$&\cellcolor[HTML]{C0C0C0}\textbf{99.2}&\cellcolor[HTML]{C0C0C0}\textbf{94.6}&\cellcolor[HTML]{C0C0C0}\textbf{96.1}&\cellcolor[HTML]{C0C0C0}\textbf{96.6}$_{(\pm0.7)}$&\cellcolor[HTML]{C0C0C0}\textbf{90.0}&\cellcolor[HTML]{C0C0C0}\textbf{65.8}&\cellcolor[HTML]{C0C0C0}\textbf{91.0}&\cellcolor[HTML]{C0C0C0}\textbf{82.3}$_{(\pm0.7)}$&\cellcolor[HTML]{C0C0C0}\textbf{92.3}&\cellcolor[HTML]{C0C0C0}\textbf{63.5}&\cellcolor[HTML]{C0C0C0}\textbf{89.7}&\cellcolor[HTML]{C0C0C0}\textbf{81.8}$_{(\pm0.9)}$ \\
\cmidrule{2-15}

&\multirow{14}{*}{100}&FT&8.0&7.5&4.7&6.7$_{(\pm0.2)}$&1.4&0.7&4.1&2.1$_{(\pm0.1)}$&2.2&1.3&1.5&1.7$_{(\pm0.1)}$\\ \cmidrule{3-15}
&&KN&0.5&0.7&0.9&0.7$_{(\pm0.1)}$&0.7&0.7&0.4&0.6$_{(\pm0.1)}$&0.3&0.8&0.4&0.5$_{(\pm0.1)}$\\
&&ROME&10.1&11.2&22.7&14.7$_{(\pm0.4)}$&34.1&22.8&68.7&41.9$_{(\pm1.2)}$&6.4&4.9&5.7&5.7$_{(\pm0.1)}$\\
&&MEMIT&1.1&1.1&1.4&1.2$_{(\pm0.1)}$&1.0&1.0&4.2&2.1$_{(\pm0.1)}$&0.6&0.9&0.7&0.7$_{(\pm0.1)}$\\
&&$\text{KE}$&0.7&0.7&0.8&0.7$_{(\pm0.1)}$&0.7&0.7&1.2&0.9$_{(\pm0.1)}$&0.8&0.8&0.7&0.8$_{(\pm0.1)}$\\
&&$\text{MEND}$&0.7&0.7&0.8&0.7$_{(\pm0.1)}$&0.7&0.7&0.8&0.7$_{(\pm0.1)}$&0.7&0.7&0.7&0.7$_{(\pm0.1)}$\\
&&$\text{MALMEN}$&54.8&52.5&66.0&57.8$_{(\pm0.9)}$&48.6&23.1&47.9&39.9$_{(\pm0.6)}$&42.1&32.4&39.2&37.9$_{(\pm0.7)}$\\
&&$\text{DAFNET}$&85.2&72.7&94.3&84.1$_{(\pm1.4)}$&73.3&42.2&77.1&64.2$_{(\pm1.1)}$&58.2&41.9&88.2&62.8$_{(\pm1.7)}$\\
&&$\text{AlphaEdit}$&76.0&60.4&81.7&72.7$_{(\pm1.0)}$&62.4&35.8&60.6&52.9$_{(\pm1.4)}$&48.5&36.9&81.6&55.7$_{(\pm0.9)}$\\ \cmidrule{3-15}
&&TP&46.6&42.0&10.4&33.0$_{(\pm0.5)}$&70.5&41.5&5.2&39.1$_{(\pm0.8)}$&45.2&29.6&12.3&29.0$_{(\pm0.7)}$\\
&&GRACE&9.8&9.2&23.7&14.2$_{(\pm0.5)}$&32.1&21.8&69.7&41.2$_{(\pm1.0)}$&6.2&5.5&5.8&5.8$_{(\pm0.2)}$\\ \cmidrule{3-15}
&&$\text{MELO}$&73.4&59.3&81.2&71.3$_{(\pm0.8)}$&54.1&32.7&57.6&48.1$_{(\pm1.2)}$&47.7&36.8&75.9&53.5$_{(\pm0.7)}$\\
&&$\text{LTE}$&78.7&63.4&83.2&75.1$_{(\pm1.6)}$&68.0&40.1&65.8&58.0$_{(\pm1.3)}$&53.9&40.2&82.6&58.9$_{(\pm1.1)}$\\
&&\cellcolor[HTML]{C0C0C0} $\text{QueueEDIT}$&\cellcolor[HTML]{C0C0C0} \textbf{89.0}&\cellcolor[HTML]{C0C0C0} \textbf{77.7}&\cellcolor[HTML]{C0C0C0} \textbf{98.1}&\cellcolor[HTML]{C0C0C0} \textbf{88.3}$_{(\pm1.0)}$&\cellcolor[HTML]{C0C0C0} \textbf{76.8}&\cellcolor[HTML]{C0C0C0} \textbf{45.7}&\cellcolor[HTML]{C0C0C0} \textbf{84.9}&\cellcolor[HTML]{C0C0C0} \textbf{69.1}$_{(\pm0.7)}$&\cellcolor[HTML]{C0C0C0} \textbf{65.4}&\cellcolor[HTML]{C0C0C0} \textbf{46.2}&\cellcolor[HTML]{C0C0C0} \textbf{90.3}&\cellcolor[HTML]{C0C0C0} \textbf{67.3}$_{(\pm0.8)}$\\ \bottomrule
\end{tabular}
\caption{Results of \texttt{QueueEDIT} and baselines with 10 and 100 edits.}
\label{edits_res_1k}
\end{table*}

\section{Implementation Details}
\label{implementation_details}
\subsection{Data}
\noindent\textbf{Training Data.} Following \cite{ZJUEditSurvey2023}, we use the ZSRE training data containing 162,555 entries, the CF training data containing 10,000 entries, and the enhanced dataset DAFSet \cite{DBLP:conf/acl/ZhangCL0HHXH24} to train meta-learning-based models, including KE \cite{KnowledgeEditor} and MEND \cite{DBLP:conf/iclr/MitchellLBFM22}.

\noindent\textbf{Evaluation Data.} We utilize three widely used datasets for evaluation. \textbf{ZSRE} \cite{DBLP:conf/conll/LevySCZ17} employs BART \cite{DBLP:conf/acl/LewisLGGMLSZ20} to answer questions followed by manual filtering, resulting in each instance containing an editing sample, a rephrased counterpart, and an irrelevant sample corresponding to reliability, generality, and locality, respectively. Inspired by \cite{ZJUEditSurvey2023}, we split the dataset into a training set and a testing set, with 162,555 and 19,009 entries. In \textbf{CF} \cite{DBLP:conf/nips/MengBAB22}, all the facts to be edited are false, thus increasing the difficulty of editing tasks. Similar to ZSRE, each data point contains an editing sample, a rephrased sample, and an irrelevant sample. Following \citet{ZJUEditSurvey2023}, both the training and testing sets comprise 10,000 entries.
\textbf{RIPE} \cite{DBLP:journals/corr/abs-2307-12976} intricately subdivides generality and locality into multiple components. Similar to CF, it involves editing false facts and is characterized by detailed evaluation. After pre-processing, a total of 4,388 entries are collected.

\noindent\textbf{General Capability Datasets.} We evaluate the general capability of various methods using these four authoritative datasets: CSQA \cite{DBLP:conf/aaai/SahaPKSC18}, MMLU \cite{DBLP:conf/iclr/HendrycksBBZMSS21}, ANLI \cite{DBLP:conf/acl/NieWDBWK20}, and SQUAD-2 \cite{DBLP:conf/emnlp/RajpurkarZLL16}. CSQA contains 200K dialogs with 1.6M turns. MMLU contains 15,908 questions. ANLI contains 3,200 test samples, and SQUAD-2 collects 8,862 questions.

\begin{figure*}[]
  \centering
  \includegraphics[width=14cm,height=7.5cm]{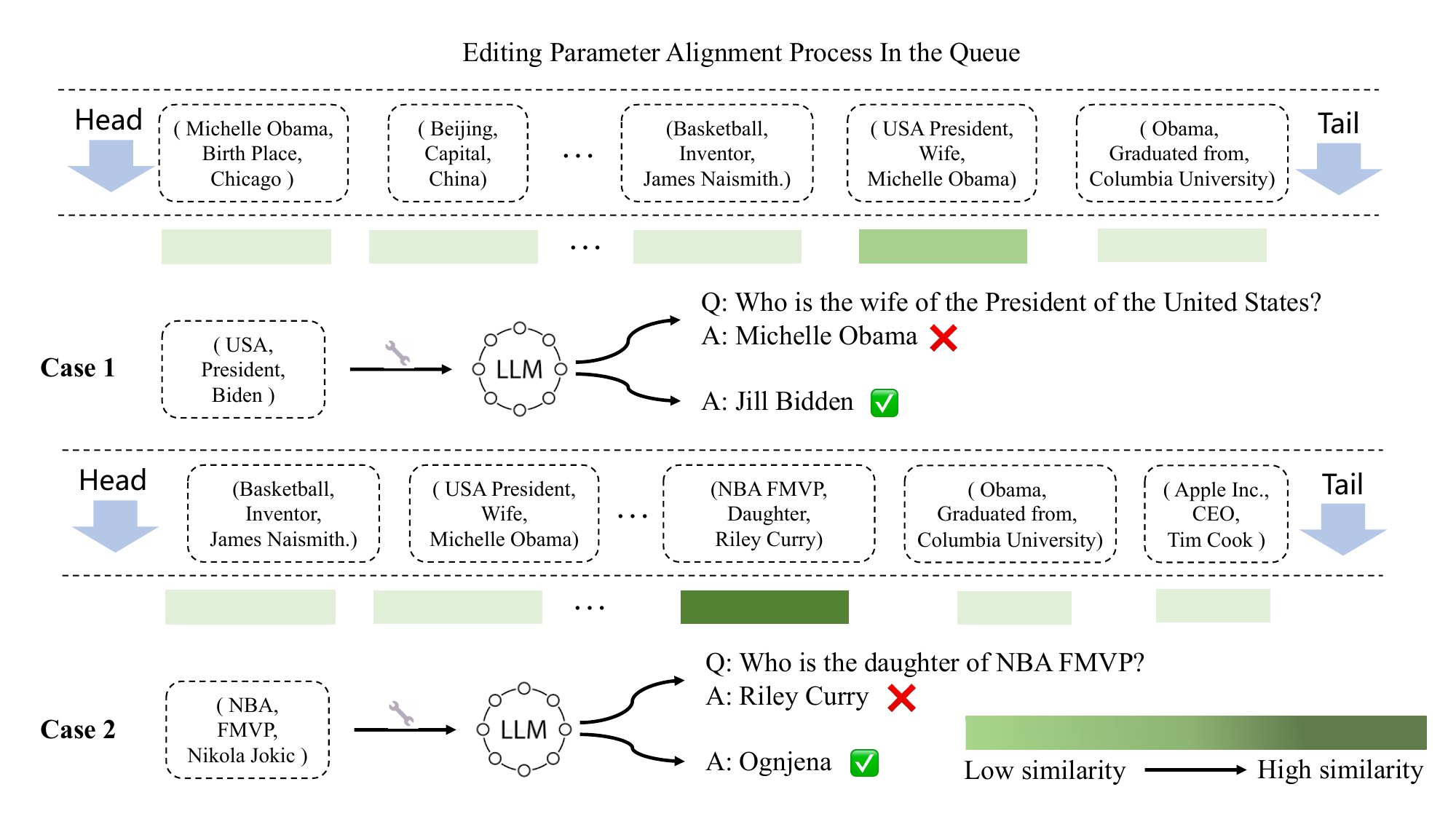}
  \vspace{-0.3cm}
  \caption{Qualitative analysis of queue-based self-correction in two case studies.}
  \label{qualitative_analysis}
  \vspace{-0.5cm}
\end{figure*}

\subsection{Baselines}
\noindent\textbf{Modify Parameters}: (1) KN \cite{DBLP:conf/acl/DaiDHSCW22} uses an integral gradient-based method to locate neurons in the FFN, achieving editing by amplifying the activation of the located neurons. 
(2) ROME \cite{DBLP:conf/nips/MengBAB22} first uses causal mediation analysis to locate the layer that has the greatest impact on the editing sample. They propose ROME to modify the FFN weights of the located layer. 
(3) MEMIT \cite{DBLP:conf/iclr/MengSABB23} expands the editing scope to multiple layers based on ROME, thereby improving editing performance and supporting batch editing.
(4) AlphaEdit \cite{DBLP:conf/iclr/FangJWMSW0C25} maps perturbations into the null space of retained knowledge prior to their integration into the parameters.
For the latter, (1) KE \cite{KnowledgeEditor} trains a bidirectional LSTM auxiliary network to predict weight updates of the editing samples. 
(2) MEND \cite{DBLP:conf/iclr/MitchellLBFM22} trains an MLP to transform the low-rank decomposition of the gradients of the model to be edited with respect to the editing samples and updates the model with the transformed gradients to achieve editing. 
(3) MALMEN \cite{DBLP:journals/corr/abs-2311-04661} accommodates editing of multiple facts with limited memory budgets by separating the computation on the hyper-network and LM, enabling an arbitrary batch size on both neural networks. 
(4) DAFNet \cite{DBLP:conf/acl/ZhangCL0HHXH24} proposes intra-inter editing attention to enhance the sequential editing of samples. In addition, the DAFSet dataset is also proposed to train the meta-learning-based editors.For baseline implementation, we use the same settings in EasyEdit \cite{DBLP:journals/corr/abs-2308-07269} for training and evaluation.

\noindent\textbf{Adding Extra Parameters}: T-Patcher \cite{T-Patcher} attaches and trains additional neurons in the FFN of the last layer of the model to be edited. GRACE \cite{GRACE} proposes General Retrieval Adapters for Continuous Editing (GRACE), which maintain a dictionary-like structure to construct new mappings for potential representations that need to be modified.

\noindent\textbf{Retrieve Data}: MELO~\cite{MELO} introduces an implementation for batch editing that utilizes LoRA technology ~\cite{DBLP:conf/iclr/HuSWALWWC22}. Meanwhile, LTE~\cite{LTE} employs fine-tuning of the LLMs to produce contextually appropriate responses when presented with knowledge preceded by specific editing cues, while simultaneously exploiting the pre-trained backbone architecture for the retrieval of relevant content~\cite{DBLP:conf/emnlp/ReimersG19}.

\subsection{Experimental Settings}

We use GPT-J\footnote{\url{https://huggingface.co/EleutherAI/gpt-j-6b}} and LLaMA3~\cite{DBLP:journals/corr/abs-2407-21783} as our backbones for editing. The queue similarity thresholds $\eta_{que}$ and $\eta_{deq}$ are set to 0.5. The hyperparameter $K$ is set to 50 due to limitations of machine resources. Regarding the selection of editing weights, we use settings consistent with MEND \cite{DBLP:conf/iclr/MitchellLBFM22} and KE \cite{KnowledgeEditor}: both GPT-J and LLaMA3 use the FFN weights of the last three layers of the model.

The upper limit for the number of sequential editing models is 1000, i.e., $T_{max}=1000$. When the number of sequential editing models reaches the maximum value, we perform an additional 20,000 iterations before stopping. We store checkpoints every 1000 iterations, and the checkpoint with the lowest loss is selected for evaluation. The learning rate $\eta$ is set to 1e-6. The training process takes 2 days on 8 NVIDIA A800 GPUs. These experiments are presented as averages from 5 random runs with different random seeds and the same hyperparameters.

\section{Editing Results}
\subsection{Results for Other Editing Quantities}
\label{apendix_seq_res}
The experimental results for 1, 10, and 100 edits are shown in Tables~\ref{single_turn_main_exp} and~\ref{edits_res_1k}, respectively. These results are consistent with our main experiments, further demonstrating the effectiveness of our approach.

\subsection{Case Study}
\label{case_study}
We further conduct a detailed analysis of the queue memory by selecting parameter blocks updated from the queue.  
Specifically, we perform a semantic similarity analysis on the top-3 most relevant triples in sequential editing knowledge from the ZSRE dataset.

\begin{itemize}
    \item \texttt{Case 1}:  
    Q: Who was the designer of Lahti Town Hall?  
    A: Eliel Saarinen.\\
    \texttt{Top-1}: Q: By which person was Lahti Town Hall designed?  
    A: Eliel Saarinen.\\
    \texttt{Top-2}: Q: When was Lahti Town Hall designed?  
    A: 1911.\\
    \texttt{Top-3}: Q: Where did the dark bricks come from for the materials used in Lahti Town Hall?  
    A: Sweden.
    
    \item \texttt{Case 2}:  
    Q: Which was the manufacturer of USS Leedstown (APA-56)?  
    A: Bethlehem Steel.\\
    \texttt{Top-1}: Q: Which corporation created USS Leedstown (APA-56)?  
    A: Bethlehem Steel.\\
    \texttt{Top-2}: Q: When was USS Leedstown (APA-56) scrapped?  
    A: 1970.\\
    \texttt{Top-3}: Q: How long did USS Leedstown (APA-56) serve?  
    A: 1943-1946.
    
    \item \texttt{Case 3}:  
    Q: In what language are the Garowe Principles written?  
    A: Somali.\\
    \texttt{Top-1}: Q: In which language is the Garowe Principles monthly football magazine reported?  
    A: Somali.\\
    \texttt{Top-2}: Q: Where were the Garowe Principles signed?  
    A: Garowe.\\
    \texttt{Top-3}: Q: Which representatives voted on The Garowe Principles?  
    A: Transitional Federal Government.
\end{itemize}

We observe that:  
(1) The MLP parameters requiring editing alignment for \texttt{Top-1} are likely semantically consistent with the current edited data.  
(2) The semantic alignment parameters for the remaining data (\texttt{Top-2} and \texttt{Top-3}) are generally consistent with the subject entities or relations.  
In summary, the top-\(K\) data aligned with the current editing knowledge in the queue is semantically consistent at the level of important entities or relations.

\subsection{Queue-Based Self-Correction Cases} 
\label{queue_self_corr}
To analyze how our \texttt{QueueEDIT} method improves the effectiveness of SME, we visualize samples from the test dataset.  
As shown in Figure~\ref{qualitative_analysis}, when asked the question, ``Who is the wife of the President of the United States?'', the strongest baseline model incorrectly answers ``Michelle Obama''.  
In contrast, our \texttt{QueueEDIT} model calculates the semantic similarity among all edited knowledge parameters in the queue and identifies that the knowledge fact ``(USA President, Wife, Michelle Obama)'' requires alignment with the current editing fact ``(USA President, Biden)'' due to high similarity.

In Case 2, our \texttt{QueueEDIT} model correctly recognizes the fact ``(NBA FMVP, Daughter, Riley Curry)'' as having the highest similarity and requiring alignment, since the NBA FMVP recipient is ``Nikola Jokic''.  
This demonstrates that our queue-based dynamic alignment mechanism can better model relationships between sequential editing data and promptly update correlations among facts.

\section{Detailed Calculation of \(v_*\)}
\label{kv_calculation}
ROME~\cite{DBLP:conf/nips/MengBAB22} aims to choose a token vector value \(v_*\) that encodes the new relation \((r, o^*)\) as a property of \(s\).  
They set \(v_* = \arg\min_z \mathcal{L}(z)\), where \(z\) is the hidden representation of \(o^*\), and the objective \(\mathcal{L}(z)\) is defined as:
\begin{equation}
\begin{split}
\mathcal{L}_z = \frac{1}{N} \sum_{j=1}^{N} & \underbrace{-\log \mathbb{P}_{G\left(m_i^{(l^*)} := z\right)}\left[o^* \mid x_j + p\right]}_{\text{(a) Maximizing } o^* \text{ probability}} \\
& + \underbrace{D_{KL}\left(\mathbb{P}_{G\left(m_{i'}^{(l^*)} := z\right)}\left[x \mid p'\right] \parallel \mathbb{P}_G\left[x \mid p'\right]\right)}_{\text{(b) Controlling semantic drift}},
\end{split}
\label{v_equation}
\end{equation}
where the first term searches for a vector \(z\) such that, when it replaces the MLP’s output at the \(i\)-th token (the subject’s final position, denoted as \(G\left(m_i^{(l^*)} := z\right)\)), the network predicts the desired object \(o^*\) in response to the factual prompt \(p\).  
The second term reduces the KL divergence between predictions for the prompt \(p'\) (of the form ``\{subject\} is a'') and those of the original model, ensuring the model retains its grasp of the subject’s core meaning.

\end{document}